\documentclass[runningheads]{llncs}

\usepackage[dvipdfmx]{graphicx}
\usepackage{multirow}
\usepackage{lscape}
\usepackage{siunitx}
\usepackage{amsmath}
\usepackage{wrapfig}
\usepackage[misc]{ifsym}
\usepackage[dvipdfmx]{graphicx}
\usepackage{mathtools}
\usepackage{amsmath,amssymb}
\usepackage{multirow}
\usepackage{xr}
\usepackage[hang,small,bf]{caption}
\usepackage[subrefformat=parens]{subcaption}
\newcommand{\Tref}[1]{Table~\ref{#1}}
\newcommand{\tref}[1]{Tab.~\ref{#1}}
\newcommand{\Eref}[1]{Equation~(\ref{#1})}
\newcommand{\eref}[1]{Eq.~(\ref{#1})}
\newcommand{\Fref}[1]{Figure~\ref{#1}}
\newcommand{\fref}[1]{Fig.~\ref{#1}}
\newcommand{\Sref}[1]{Section~\ref{#1}}
\newcommand{\sref}[1]{Sec.~\ref{#1}}

\begin{document}

\title{Can Transformers Predict Vibrations?}

\titlerunning{Can Transformers Predict Vibrations?}

\author{Fusataka Kuniyoshi \and Yoshihide Sawada}
\institute{AISIN Corporation, Aichi, Japan}

\toctitle{Can Transformers Predict Vibrations?}
\tocauthor{F. Kuniyoshi, Y. Sawada}

\maketitle
\begin{abstract}
Highly accurate time-series vibration prediction is an important research issue for Electric vehicles~(EVs). EVs often experience vibrations when driving on rough terrains, known as torsional resonance. This resonance, caused by the interaction between motor and tire vibrations, puts excessive loads on the vehicle's drive shaft. However, current damping technologies only detect resonance after the vibration amplitude of the drive shaft torque reaches a certain threshold, leading to significant loads on the shaft at the time of detection. In this study, we propose a novel approach to address this issue by introducing Resoformer, a transformer-based model for predicting torsional resonance. Resoformer utilizes time-series of the motor rotation speed as input and predicts the amplitude of torsional vibration at a specified quantile occurring in the shaft after the input series. By calculating the attention between recursive and convolutional features extracted from the measured data points, Resoformer improves the accuracy of vibration forecasting. To evaluate the model, we use a vibration dataset called VIBES (Dataset for Forecasting Vibration Transition in EVs), consisting of 2,600 simulator-generated vibration sequences. Our experiments, conducted on strong baselines built on the VIBES dataset, demonstrate that Resoformer achieves state-of-the-art results. In conclusion, our study answers the question ``Can Transformers Forecast Vibrations?'' While traditional transformer architectures show low performance in forecasting torsional resonance waves, our findings indicate that combining recurrent neural network and temporal convolutional network using the transformer architecture improves the accuracy of long-term vibration forecasting.

\keywords{Electric Vehicles (EVs) \and Time-series Forecasting \and Torsional Resonance \and Transformer}
\end{abstract}
\section{Introduction} 
\label{sec:introduction}
Highly accurate time-series forecasting is a crucial technology in numerous fields and has been extensively studied. However, in niche domains like Electric vehicles~(EVs) motor control, the utilization of state-of-the-art forecasting technology remains untapped. This paper aims to address this gap by exploring the application of AI in this specific area and making a significant contribution to its development.
 
EVs are increasingly being recognized as a crucial solution to address the current fossil-energy crisis and mitigate climate change by reducing carbon emissions. Their widespread adoption offers numerous benefits both for local and global environments, making them a promising option for sustainable transportation~\cite{Li2015HiddenBO}.

The rapid advancements in EV technology have resulted in an exponential increase in demand for electric powertrains. Effective damping control is crucial to ensure the safe operation and optimal performance of electric powertrains. Numerous solutions have been proposed to address the problem of torsional resonance in the development of EVs, PD controller design~\cite{Karamuk2019DesignOA}, improved anti-jerk controller performance~\cite{Scamarcio2020ComparisonOA}, and torsional vibration analysis~\cite{Tang2017ANS}.

\begin{figure}[t]
  \begin{minipage}[t]{0.5\linewidth}
    \centering
    \includegraphics[keepaspectratio, scale=0.95]{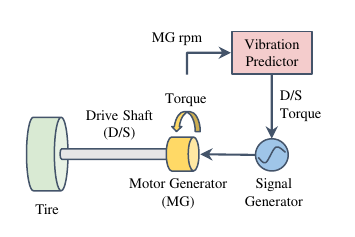}
    \subcaption{}
    \label{fig:scheme}
  \end{minipage}
  \begin{minipage}[t]{0.5\linewidth}
    \centering
    \includegraphics[keepaspectratio, scale=0.95]{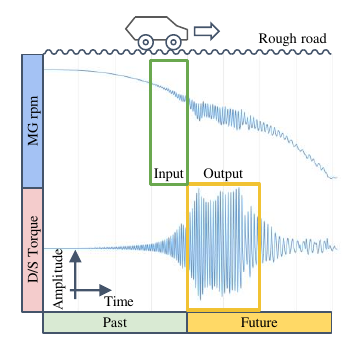}
    \subcaption{}
    \label{fig:diagram}
  \end{minipage}
  \caption{Schematic representation of our proposed pipeline for forecasting vibration in EVs. (a) The overall scheme illustrating the vibration reduction process. (b) Diagram of vibration transition in EVs when driven on rough roads that highlights the relationship between past MG rpm inputs and predicted future D/S torque outputs.}
  \label{fig:overview}
\end{figure}

In the EV drivetrain, the torsional resonance refers to the ``twisting'' movement of the rotating shafts that connect the motor generator (MG) and tires or various components of the drivetrain. In an electric-powertrain system, torsional resonance occurs in the drive shaft (D/S) that connects the MG and tire, as illustrated in~\Fref{fig:scheme}. Torsional resonance, which is characterized by large and destructive amplitude fluctuations, poses a significant risk to the integrity of the D/S of a vehicle. The excessive load that is generated by such resonance can result in D/S failure, compromised vehicle performance, and even hazardous driving conditions; therefore, effective sequential detection and control measures are necessary.

The use of machine learning-based sequential modeling is a promising approach that has gained attention in computer science, natural science, and industrial technology. This approach is widely used in various applications, including machine translation~\cite{Sutskever2014SequenceTS,Vaswani2017AttentionIA}, the extraction of meaningful patterns from text~\cite{Huang2015BidirectionalLM,Zhang2015CharacterlevelCN}, and audio processing~\cite{Oord2016WaveNetAG,GRAVES2005602}. The application of machine learning in the electric powertrain control field is highly anticipated, with significant progress in the development of electric motors and vehicle systems. Several studies have already been published in this field, such as those on remaining-range prediction~\cite{Zhao2020MachineLM,Kim2022AML}, energy-consumption estimation modeling~\cite{Shen2022ElectricVE}, vehicle-component temperature estimation~\cite{Kirchgssner2020DataDrivenPM,Kirchgssner2021EstimatingEM,rustamov2022lifenet,warey2020electric}, noise estimation~\cite{Tan2020AML,Qian2020SoundQE}, and road-profile or driving-style prediction~\cite{Park2009IntelligentVP,Ghadge2015MachineLA,Alrefai2022InVehicleDF}.

Vibration control is a fundamental technology in various applications; however, current techniques predominantly rely on the straightforward incorporation of machine-learning methods~\cite{Yang2022TimeSF,Xia2022ResearchOF,Zollanvari2021TransformerFP} or controller design~\cite{Lin2018AcceleratedAS,Karamuk2019DesignOA,9709676,Zhang2020ActiveOC}; the exploration of deep learning approaches remains limited. Moreover, few studies have investigated deep-learning approaches for mitigating the shaft load in EVs. Therefore, we propose a cutting-edge attention-based deep learning model for data-driven time-series analysis to predict vibration transitions.

The conventional solution for damping torsional resonance requires the installation of a torque-detection sensor (observer) on the driving shaft to measure the D/S-torque, which increases the system complexity and cost. To overcome this limitation, we propose a novel scheme that incorporates a vibration predictor that can predict the future D/S-torque from the MG rotation per minute (MG rpm). This enables the inference of the future D/S-torque without past D/S-torque data being required. \Fref{fig:scheme} depicts a scheme in which the future torque of the D/S is predicted using a vibration predictor based on time-series data of the MG rpm. The MG is subsequently controlled by a signal generator that produces a control signal that is derived from the predicted value. Consequently, accurate prediction of the D/S-torque amplitude is crucial to generate effective control signals that mitigate excessive loads on the D/S. Damping control is performed based on the predicted torque value to output the torque to the motor in the opposite phase to the torque vibration that is transmitted to the D/S. Furthermore, electric-motor control is recognized for its rapid response, from torque command to torque output~\cite{Hori1999MotionCO,Li2018HierarchicalBT}, as well as its precise determination of the output-torque magnitude, which enables efficient and accurate performance in various applications. Our configuration can predict the future value of the torque that is transmitted to the D/S without using the detected torque value. Instead, machine learning is used for the prediction based on the detected value of the rotational speed of the motor. Therefore, no torque-detection sensor is required and the configuration of the vehicle controller can be simplified.

This study uses machine-learning techniques to predict the torsional-resonance waveforms that arise during rough road driving. A schematic of the significant resonance that occurs during such driving is depicted in~\fref{fig:overview}. The yellow rectangle in the figure illustrates the specific section in which large resonance occurs following braking.

In summary, our aim is to forecast future time-series data and to develop generic machine-learning models to predict vibration transitions in EVs. The contributions of this study include the following:
\renewcommand{\labelenumi}{(\arabic{enumi})}
\begin{enumerate}
    \item We propose a novel scheme for forecasting vibration transition that can predict the future D/S-torque from the MG rpm.
    \item We propose an attention-based Transformer architecture named Resoformer to forecast the vibration transition by observing the amplitude of the torsional resonance that achieves high performance on VIBES. 
    \item We conduct recent transformer-based time-series forecaster on a case of torsional resonance in EVs.
\end{enumerate}

We implement various strategies to tailor the architecture to surpass the performance of existing state-of-the-art benchmarks. We ensure that it can handle the necessary inputs for vibration forecasting tasks by specifically incorporating the following: (1) recurrent and convolutional layers that encode time-variant features and context vectors for use in other parts of the network; (2) an attention layer to identify and process long-term dependencies that exist in the input sequence; and (3) gating mechanisms and sample dependence to minimize the contributions of irrelevant inputs. The use of these specialized components also facilitates the experiments. In particular, we evaluate the performance of our method by demonstrating the detailed process of training the prediction model with sufficient evaluations. We also analyze the proposed method~(Resoformer) using real-world torsional-resonance data on our simulation-data trained model. The overall pipeline is shown in~\Fref{fig:overview}.

The remainder of this paper is organized as follows. In \Sref{sec:relatedwork}, we provide a review of related work in the field. \Sref{sec:blocks} presents an explanation of the fundamental building blocks of our proposed model architecture. In \Sref{sec:resoformer}, we introduce the Resoformer, our transformer-based architecture designed specifically for forecasting torsional resonance. Next, in \Sref{sec:evaluation}, we present the results obtained from our experiments and compare the accuracy of the proposed Resoformer model with the baseline models. We also provide insights based on our dataset. Finally, in \Sref{sec:conclusion}, we offer concluding remarks and discuss potential future directions for research.

\section{Related Work}
\label{sec:relatedwork}
Time-series forecasting using deep neural networks (DNNs) encompasses numerous architectures, many of which primarily focus on autoregressive or iterated multi-step forecasting techniques~\cite{Ariyo2014StockPP,Bahdanau2014NeuralMT,SALINAS20201181,Taylor2018ForecastingAS,Rangapuram2018DeepSS}. However, these approaches tend to exhibit a significant accumulation of errors in long-term forecasting scenarios~\cite{Zeng2022AreTE}.

Several Transformer-based models have been proposed to address the challenges that are associated with long-term time-series forecasting~\cite{Vaswani2017AttentionIA,Wu2021AutoformerDT,Zeng2022AreTE,Liu2022NonstationaryTE,Liu2023iTransformerIT,Zhang2023CrossformerTU,Nie2022ATS}. These models relay on encoder--decoder structures, which can result in a slow inference speeds when the output sequence is long. Thus, with in-vehicle in mind, we opt to forgo Transformer-based encoder--decoder models in our approach.

DNN architectures such as the recurrent neural network (RNN)~\cite{Dauphin2016LanguageMW,Bai2018AnEE,HEWAMALAGE2021388} have been used extensively for time-series forecasting applications. These methods include the incorporation of attention layers into the RNN outputs~\cite{Wang2019RTransformerRN,LIM20211748,9645961,You2019HierarchicalTC}, the input of features that are extracted by convolution into the recurrent layer~\cite{Lai2017ModelingLA,Jin2019MultistepDA}, the minimization of both shape and temporal loss~\cite{Guen2019ShapeAT}, and the use of quantile regressors based on the RNN~\cite{Wen2017AMQ,Kan2022MultivariateQF}. Such methods generally rely on either an RNN or a convolutional neural network (CNN) layer as the foundation. Instead, our proposed methodology incorporates a parallel RNN and CNN architecture, which integrates and fuses features from both models to improve the prediction accuracy.

\section{Building blocks} 
\label{sec:blocks}
Our proposed model architecture, Resoformer, belongs to a recent class of sequence models for deep learning that combines elements from RNN, temporal convolutional network (TCN), and gating mechanisms. Resoformer aims to leverage the strengths of both RNN and TCN, drawing inspiration from convolutional sequence models~\cite{Gehring2017ConvolutionalST}. However, there are several key differences. Instead of using attention, Resoformer incorporates a transformer block, and it employs two different model architectures (RNN and TCN) in parallel.

In this section, we first introduce our task in~\Sref{sec:task}. Then, we provide a review of the two sequence modeling methods that serve as the building blocks of our model. Specifically, we discuss the RNN architecture for extracting conditional temporal dynamics in~\Sref{sec:rnn}, and the TCN for extracting convolutional features in~\Sref{sec:tcn}.

\subsection{Task Definition}
\label{sec:task}

The input sequence consists of $L$ data steps for a time-series sequence, which is represented as $\mathbf{X}=\{x_0, \cdots, x_L\}$. The aim is to predict the values in the next $T$ time steps, which are represented as $\mathbf{Y}=\{y_{L+1}, \cdots, y_{L+T}\}$. The input sequence $\mathbf{X}$ corresponds to the MG rpm and the predicted sequence $\mathbf{Y}$ corresponds to the D/S torque.

Vibrational transition forecasting is conceptualized as a supervised machine learning task. The input length $k$ at time $t$ is denoted as $\mathcal{X}_{t-1-k:t-1}=\{x_{t-1-k}, \cdots, x_{t-1}\}$, encapsulating all historical information within a predefined look-back window $k$. The aim is to predict the output signal $\mathcal{Y}$ over the next $\tau$ time steps, formulated as $\mathcal{Y}_{t:t+\tau}=\{y_{t}, \cdots, y_{t+\tau}\}$. To this end, the proposed approach adopts quantile regression. Each quantile forecast utilizes $\tau$-step-ahead forecasting models at time $t$, expressed as follows.

\begin{equation}
    \{ \hat{y}_{t} \} = f(\tau, x_{t-1-k}, \cdots, x_{t-1}),
\end{equation}

where $f(\cdot)$ is the model deployed to obtain the predictive output.

In a typical time-series forecasting task, we consider pairs of signals from the same domain. In this case, we chose to set $\mathcal{X}$ as the rpm of the MG and $\mathcal{Y}$ as the torque in D/S.

\subsection{Recurrent Neural Network}
\label{sec:rnn}
The RNN estimates an effective distribution $P(\textbf{x}_{0:n})$ of the sequences of data points $\mathbf{x} = \{x_0,x_1,\dots,x_n\} \eqqcolon \textbf{x}_{0:n}$, where $n$ is the serial number of data points. In long-short-term memory (LSTM), one type of a RNN~\cite{Hochreiter1997LongSM}, the computation is carried out recursively with the assistance of an additional recurrent quantity $\mathbf{c}_t$, which is known as a memory cell.

\begin{eqnarray}
    \mathbf{h}_n(\mathbf{x}_{0:n-1}) = f_h(\mathbf{x}_{n-1}, \mathbf{h}_{n-1}, \mathbf{c}_{t-1}; \theta),\label{eq:rnn}
\end{eqnarray}

where $\theta$ refers to all network parameters. The initial values of $\mathbf{h}_{n-1}$ and $\mathbf{c}_{n-1}$ can be set to zero or treated as part of the model parameters $\theta$.

\subsection{Temporal Convolutional Network}
\label{sec:tcn}
The TCN consists of one-dimensional stacked convolutional layers that enable the efficient and effective processing of temporal data~\cite{Bai2018AnEE}. Specifically, when a length $n$ sequence $\mathbf{x}$ is fed into the model, the convolutional transformation $\mathcal{H}(\cdot)$ is expressed as a function of the input and a set of trainable parameters. This results in a TCN output $\mathbf{u}$ that represents a series of transformations in the input sequence $\mathbf{x}$.

\begin{eqnarray}
    \mathcal{H}(\mathbf{x}_t) &=& \sum_{i=0}^{k}f(i)\cdot\mathbf{x}(t-(l \cdot i)),\\
    \mathbf{u} &=& [\mathcal{H}(\mathbf{x}_1),\mathcal{H}(\mathbf{x}_2),\dots,\mathcal{H}(\mathbf{x}_n)],\label{eq:tcn}
\end{eqnarray}

where $x_t$ is the input at time step $t$, $f$ is the filter with size $k$, $l=2^v$ is the dilation factor, and $v$ is the network depth.

\subsection{Dot-Product Attention}
The Transformer architecture~\cite{Vaswani2017AttentionIA} employs an attention mechanism to compute representations of input sequences. The attention mechanism that is used in the Transformer is known as the scaled dot-product attention, which operates on three sets of vectors: queries $\mathbf{Q}$, keys $\mathbf{K}$, and values $\mathbf{V}$.

\begin{eqnarray}
    \mathrm{head_i} = \mathrm{Softmax}\left( \frac{\mathbf{Q} \mathbf{K}^T}{\sqrt{d_k}} \right) \mathbf{V},
\end{eqnarray}

where $d_k$ is the dimension of the keys and $\sqrt{d_k}$ is the scaling factor. This is used for parallel attention computations to produce output values that are concatenated as follows:

\begin{eqnarray}
    \mathrm{Att}(\mathbf{Q},\mathbf{K},\mathbf{V}) = [\mathrm{head}_1; \mathrm{head}_2; \dots; \mathrm{head}_h]\mathbf{W}_o,
\end{eqnarray}

where $h$ denotes the number of heads and $\mathbf{W}_o$ is a linear projection to obtain the final output value. This method is known as multi-head attention, and it enables the model to focus on different aspects of the input simultaneously and to capture more complex patterns.

\subsection{Gating Mechanism}
It can be challenging to anticipate which variables are relevant because the precise relationship between inputs and targets is typically unknown in advance. Therefore, we introduce component gating layers based on gated linear units (GLUs)~\cite{Dauphin2016LanguageMW}. GLUs provide our model with the flexibility to suppress any parts of the architecture that may not be useful when capturing the relevant features. The GLUs use full inputs as their respective self-gates and compute the element-wise product upon themselves.

\begin{eqnarray}
    T(\mathbf{X}) &=& \sigma(\mathbf{X}\mathbf{W}_1 + \mathbf{b}_1),\\
    \mathrm{GLU}(\mathbf{X}) &=& T(\mathbf{X}) \otimes (\mathbf{X}\mathbf{W}_2 + \mathbf{b}_2),
\end{eqnarray}

where $T(\mathbf{X})$ indicates the transform gate, $\otimes$ is the element-wise Hadamard product, $\sigma(\cdot)$ is the activation function of the sigmoid, and $\mathbf{W}_1, \mathbf{W}_2$, and $\mathbf{b}_1, \mathbf{b}_2$ are model parameters.

\section{Resoformer for Predicting Vibration in EVs}
\label{sec:resoformer}
Improving the metrics against the popularity baseline is not a trivial task. We rely on a large dataset of torsional resonance which includes 2,600 driving sequence. Additionally, we use a custom transformer architecture which captures the interactions between those features and models the problem by taking into account the future time-series forecasting.

We design a novel architecture that uses canonical components to build feature representations for high prediction performance in a torsional resonance problem. The model combines features from LSTM and TCN networks, learns embeddings for whole input series, and concatenates everything into a dense layer which is followed by two hidden layers leading to a output layer. Specifically the model configuration is the following:

\renewcommand{\labelitemi}{$\bullet$}
\begin{quote}
    \begin{itemize}
        \item Input embedding strategies: We adopt two sequence modeling methods, LSTM and TCN, to extract time-variant characteristics from the input sequence (\Sref{sec:encoding}).
        \item Attention schemes: Transformers rely on the attention mechanism to extract semantic dependencies between paired sequences (\Sref{sec:attention}).
        \item Gating mechanism: This mechanism is used to skip over any unused components of the architecture, thereby providing adaptive depth and network complexity to accommodate a wide range of scenarios (\Sref{sec:gating}).
    \end{itemize}
\end{quote}

\begin{figure}[t]
\centering
    \includegraphics[width=0.8\linewidth]{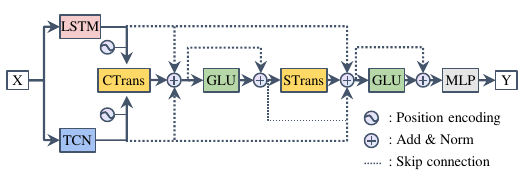}
    \caption{Conceptual diagram of our proposed method.}
  \label{fig:architecture}
\centering
\end{figure}

Our proposed model architecture is depicted in~\fref{fig:architecture}. The output is fed into mean absolute loss (MAE).

\subsection{Input Sequence Encoding and Position Encoding}
\label{sec:encoding}
Unlike LSTM or RNN, the original Transformer has no recurrence. Instead, it uses the positional encoding that is added to the input embeddings to model the sequence information. We summarize some positional encodings ($\mathrm{PE}$), which are added element-wise to the dense embeddings of the input data points that are represented by $\mathbf{E}\in\mathbb{R}^{L\times dh}$ where $L$ is the sequence length, $d$ is the input dimension of each head, and $h$ is the number of attention heads, before being input to the Transformer modules:

\begin{eqnarray}
    \mathrm{PE}(pos, 2i) &=& \sin\left(\frac{pos}{10000^{2i/d}}\right),\\
    \mathrm{PE}(pos, 2i+1) &=& \cos\left(\frac{pos}{10000^{2i/d}}\right).
\end{eqnarray}

The positional index ``$pos$'' is an embedding of the sequence position, where $\sin(\cdot)$ and $\cos(\cdot)$ are the hand-crafted frequencies for each dimension~$i$.

\begin{eqnarray}
    \mathbf{X}_{rnn} &=& \mathrm{RNN}(\mathbf{E}) + \mathrm{PE}(\mathbf{E}),\\
    \mathbf{X}_{tcn} &=& \mathrm{TCN}(\mathbf{E}) + \mathrm{PE}(\mathbf{E}).
\end{eqnarray}

We incorporate $\mathrm{PE}$ into both the RNN and TCN models, which is denoted as $\mathbf{X}_{rnn}$ and $\mathbf{X}_{tcn}$ in \Eref{eq:rnn} and \eref{eq:tcn}, respectively. This approach enhances the ability of the model to capture temporal and position information within the input sequences.

\subsection{Transformer Block}
\label{sec:attention}
Given input representations~$\mathbf{X}$, the Transformer components undergo a sternward layer normalization ($\mathrm{LN}$), and the Transformer output is denoted by $\mathbf{O}\in \mathbb{R}^{L\times dh}$:

\begin{eqnarray}
    \mathbf{A} &=& \mathrm{LN}(\mathbf{X}+\mathrm{Att}(\mathbf{Q}, \mathbf{K}, \mathbf{V})),\\
    \mathrm{FFN} &=& \mathrm{FF}(\mathrm{ReLU}(\mathrm{FF}(\mathbf{A}))),\\
    \mathbf{O} &=& \mathrm{LN}(\mathbf{A}+\mathrm{FFN}(\mathbf{A})),\label{eq:transformer}
\end{eqnarray}

where $\mathrm{FFN}$ are the position-wise feedforward networks, $\mathrm{FF}$ denotes the feedforward fully connected layer, $\mathbf{A}$ is the intermediate representation, and $\mathrm{ReLU}$ is the nonlinear activation function.

\subsection{Gating and Skip Connection}
\label{sec:gating}
In Resoformer, we utilize multi-head pairwise attention to handle pairs of sequence data. However, this approach may not fully capture the significance of individual features. To partially overcome this limitation, we incorporate gating into the Transformer architecture by introducing a new input branch. This idea is inspired by the work of Chai et al.~\cite{Chai2020HighwayTS}. This enriches the representations of the residual connection with augmented gating-modified encoding. We also add gated units to the $\mathrm{FFN}$ modules to provide an additional self-adaptive information flow. From another perspective, the GLU can be considered as a self-dependent nonlinear activation function with dynamic adaptation. The self-gating augmented Transformer module is calculated as follows:

\begin{eqnarray}
    \mathbf{U} &=& \mathrm{LN}(\mathrm{CTrans}(\mathbf{X}_{rnn}, \mathbf{X}_{tcn}) + \mathbf{X}_{rnn} + \mathbf{X}_{tcn}),\\
    \mathbf{V} &=& \mathrm{LN}(\mathbf{U} + \mathrm{GLU}(\mathbf{U})),\\
    \mathbf{W} &=& \mathrm{LN}(\mathrm{STrans}(\mathbf{V}) + \mathbf{V} + \mathbf{X}_{rnn} + \mathbf{X}_{tcn}),\\
    \mathbf{O} &=& \mathrm{LN}(\mathbf{W} + \mathrm{GLU}(\mathbf{W})),\\
    \mathbf{Y} &=& \mathrm{MLP}(\mathbf{O}).
\end{eqnarray}

In this formulation, $\mathbf{U}$, $\mathbf{V}$, $\mathbf{W}$, and $\mathbf{O}$ denote intermediate representations, whereas $\mathbf{Y}$ represents output. $\mathrm{CTrans}(\cdot,\cdot)$ denotes a transformer that uses co-attention to compute the attention scores between two different sequences, $\mathbf{X}_{rnn}$ and $\mathbf{X}_{tcn}$. Moreover, $\mathrm{STrans}(\cdot)$ refers to a Transformer that employs self-attention to compute the attention scores within a single sequence. The Transformer equation is described in~\eref{eq:transformer}. Both the self-attention and co-attention mechanisms use the same scaled dot-product attention method, but with different input sets, which enables effective feature extraction and fusion. The final prediction is subsequently obtained through the fully connected two layers (Multi-Layer Perceptron, MLP) across the channel dimensions. The overall Resoformer diagram is presented in~\fref{fig:architecture}.

\section{Performance Evaluation}
\label{sec:evaluation}
In this section, we provide a detailed description of the dataset used to evaluate our model's performance on torsional resonance in EVs in~\Sref{sec:dataset}. We explain our training process for performance evaluation in~\Sref{sec:training}. Furthermore, we introduce the models used for performance evaluation in~\Sref{sec:models}. The results of the performance evaluation are presented in~\Sref{sec:results}. Additionally, we showcase visualization results comparing the ground truth vibration with the prediction results in~\Sref{sec:vizresults}.

\subsection{Dataset}
\label{sec:dataset}
We used a dataset to predict the occurrence of significant torsional resonance~\cite{kuniyoshi2024} that arises when there is abrupt deceleration during driving in rough terrain.

For supervised learning, we partitioned all time series into three parts: 2,000 for the training set for learning, 600 for the hold-out test set for performance evaluation. We use a fixed-length sliding time window approach~\cite{Wu2020DeepTM} to construct input $L$ and output $T$ pairs for the model training and evaluation. Input--output pairs are created by sliding a window per sample with a width of $L$ samples for the input and $T$ samples for the output in the time-axis direction for each sequence.

\subsection{Training}
\label{sec:training}
Before benchmarking, an extensive hyper-parameter tuning phase was undertaken for various models, including the Transformer, Autoformer, DLinear, NonStationary, iTransformer, Crossformer, and PatchTST. This phase involved training each model using a specified training dataset, followed by performance evaluation on a distinct test dataset over 50 epochs—deliberately set lower than the epochs for benchmarking—to concentrate on the influence of hyper-parameters. The process aimed to identify the optimal learning rate through five separate training cycles for each model, choosing the rate that consistently minimized loss. Evaluated learning rates included $10^{-5}$, $10^{-4}$, and $10^{-3}$. Additionally, minibatch sizes of 8, 16, 32, and 512 were tested to ascertain their impact on model efficacy. Cosine annealing, as proposed by Loshchilov and Hutter~\cite{Loshchilov2016SGDRSG}, was utilized for learning rate scheduling during this tuning stage. The performance scores recorded across the five pre training iterations. This evaluation process was facilitated by the Time Series Library (TSlib)~\cite{wu2023timesnet}\footnote{\url{https://github.com/thuml/Time-Series-Library}}, which provided a robust framework for assessing the models. Within TSlib’s configurations, the dimension of the feed-forward network ($d_{ff}$) was set to four times the model dimension ($d_{model}$), aligning with the library's recommendations. We set $d_ff = 4 \times d_model$.
Detailed outcomes of this hyper-parameter optimization of the optimized learning rates for each model are documented in Table~\ref{tab:optimized_lr}. Additionally, we apply minibach size: 512, $\texttt{d\_model}$: 128, and $\texttt{d\_ff}$: 512 to the Transformer, Autoformer, DLinear, NonStationary, iTransformer, Crossformer, and PatchTST.

For Resoformer, LSTM, and TCN, we set the mini-batch size to 1024. We used the Adam optimizer~\cite{Kingma2014AdamAM} with $b_1 = 0.9$, $b_2 = 0.999$, and $\epsilon = 10^{-7}$. The learning rate was set to $0.0005$. We applied dropout techniques for each layer: the self-attention, feedforward, and normalization sublayers. A dropout rate of 0.2 was used for each sublayer. Moreover, the dimension of the hidden layer of the LSTM and the number of filters for the TCN and hidden state of attention was set to 64. For the TCN, we set the kernel size to 3 and the dilations to $\{16, 32, 64, 128\}$.

All of the models are trained by minimizing the Mean Absolute Error (MAE) which is the median of the predictive distribution. The MAE's equation is the following.

\begin{equation}
    \text{MAE} = \frac{1}{n} \sum_{i=1}^{n} |y_i - \hat{y}_i|
\end{equation}

In this formula, $n$ represents the total number of data points, $y_i$ denotes the actual values, and $\hat{y}_i$ signifies the predicted values. The MAE is calculated by dividing the sum of the absolute errors between the actual and predicted values by the total number of data points.

\begin{table}[t]
\begin{center}
\caption{Variable and constant parameters in whole 2,600 series. We split dataset to 2,000 series for train dataset and 600 series for test dataset. The difference of two datasets is stiffness, the others parameters are same between train and test dataset. The bold parameters means these parameters are included in test dataset.}
\label{tab:parameters}
    \begin{tabular}{lc}
        Parameters & Values \\ \hline
        Stiffness [N/m] & \begin{tabular}{c} 2662.0, 3771.2, 4880.3, 5989.5, 7542.3,\\ 9095.2, 10648.0, 12644.7, 14641.3, 16638.0 \\ \textbf{1500.0, 6800.0, 18000.0} \end{tabular} \\ \hline
        Brake [s] & \begin{tabular}{c} 2.1, 2.2, 2.3, 2.4, 2.5, 2.6, 2.7, 2.8, 2.9, 3.0,\\ 4.0, 5.0, 6.0, 7.0, 8.0, 9.0, 10.0, 11.0 12.0, 13.0 \end{tabular} \\ \hline
        Road friction ($\mu$) & 0.2, 0.4, 0.6, 0.8, 1.0 \\ \hline
        Motor position & Front, rear \\
    \end{tabular}
\end{center}
\end{table}

\subsection{Benchmarking Models}
\label{sec:models}
We conducted three sets of experiments to evaluate the performance of the proposed Resoformer. The objective of the first experiment is to use simulation-generated datasets to investigate the performance of the Resoformer on VIBES, thereby establishing comparisons against Resoformer. Eleven different models are applied: Transformer~\cite{Vaswani2017AttentionIA}, Autoformer~\cite{Wu2021AutoformerDT}, DLinear~\cite{Zeng2022AreTE}, NonStationary~\cite{Liu2022NonstationaryTE}, iTransformer~\cite{Liu2023iTransformerIT}, Crossformer~\cite{Zhang2023CrossformerTU}, and PatchTST~\cite{Nie2022ATS}. Additionaly, the LSTM model~\cite{Sutskever2014SequenceTS}, which combines an LSTM with two fully connected two-layer multilayer perceptrons (MLPs); TCN~\cite{Bai2018AnEE}, which is explained in \Sref{sec:tcn}. Our analysis involves the inclusion of the Zero model—a baseline where model outputs are set to zero across the board. These baselines serve the predictive capabilities of the other models and setting a foundational benchmark for comparison.

\subsection{Results and Analysis}
\label{sec:results}
The comparative analysis of forecasting models, as detailed in~\tref{tab:mae}, sheds light on the nuanced performance dynamics across different prediction lengths, with a particular focus on MAE as the evaluation metric. This analysis is crucial for understanding the landscape of current forecasting methodologies, highlighting the effectiveness of various models, including our proposed Resoformer, in the context of time-series forecasting.

Notably, the results reveal a trend where classical approaches, such as the LSTM and TCN models, outperform Transformer-based models in terms of accuracy. This observation is particularly striking, as it suggests that, despite the advancements in neural network architectures, traditional methods retain a competitive edge in specific forecasting contexts. The LSTM model, for instance, demonstrates commendable performance, surpassing several of its more contemporary counterparts and highlighting the enduring relevance of classical neural network designs. Amidst this landscape, our proposed method, the Resoformer, distinguishes itself by achieving higher scores than the LSTM, indicating its superior capability in capturing the intricacies of time-series data. However, it falls short of surpassing the TCN model, which remains the top performer across all evaluated prediction lengths. This places the Resoformer in a unique position, bridging the gap between classical methodologies and the latest advancements in neural network architectures.

The distinction between the performance of Transformer-based models and classical approaches underscores a critical insight into the nature of time-series forecasting. It suggests that while transformer models offer significant benefits for a wide array of applications, classical methods like LSTM and TCN continue to hold significant value, especially in scenarios where their specific architectural advantages come to the fore.

Furthermore, the positioning of the Resoformer as a viable alternative that combines elements of both traditional and modern approaches reflects our intention to leverage the best of both worlds. By outperforming the LSTM and closely competing with the TCN, the Resoformer exemplifies the potential of hybrid models in advancing the state-of-the-art in time-series forecasting.

\begin{table}[t]
    \caption{Forecasting results with different prediction lengths \(T\in\{96,192,336,720\}\). We set the input length \(L\) to 192 for VIBES. The highest and second-highest for each metric are indicated in bold and underlined, respectively. The metrics used are MAE.}
    \centering
    \begin{tabular}{lccccc}
        Model & T=720 & 336 & 192 & 96 \\
        \hline
        Transformer & 0.549 & 0.438 & 0.342 & 0.262 \\
        Autoformer & 0.778 & 0.756 & 0.632 & 0.598 \\
        DLinear & 0.649 & 0.613 & 0.586 & 0.568 \\
        NonStationary & 0.566 & 0.504 & 0.405 & 0.287 \\
        iTransformer & 0.642 & 0.606 & 0.533 & 0.440 \\
        Crossformer & 0.559 & 0.454 & 0.364 & 0.289 \\
        PatchTST & 0.637 & 0.565 & 0.467 & 0.373 \\
        Zero & 0.655 & 0.618 & 0.593 & 0.575 \\
        LSTM & 0.354 & 0.330 & 0.315 & 0.306 \\
        TCN & \textbf{0.338} & \textbf{0.297} & \textbf{0.259} & \textbf{0.214} \\
        Resoformer & \underline{0.345} & \underline{0.305} & \underline{0.308} & \underline{0.226} \\
    \end{tabular}
    \label{tab:mae}
\end{table}


\subsection{Visualized Results}
\label{sec:vizresults}
Graphical depictions in our figures provide a nuanced understanding of each model's long-term forecasting capabilities, supplementing the quantitative data in~\tref{tab:mae}. These visualizations are particularly insightful for assessing the Resoformer model's performance nuances at extended forecast horizons, highlighting strengths and weaknesses not immediately obvious from statistical metrics alone.

\begin{figure}[t]
    \begin{tabular}{cc}
      \begin{minipage}[t]{0.5\hsize}
        \centering
        \includegraphics[keepaspectratio, scale=0.25]{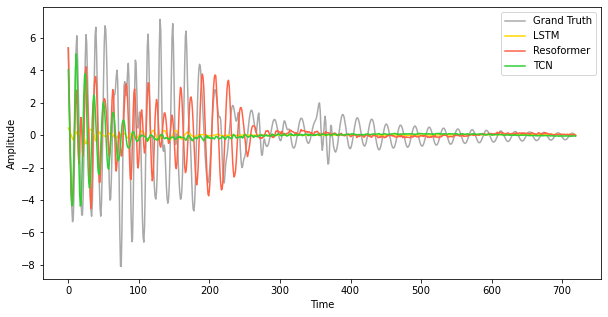} \\
        \subcaption{}
        \label{res1_a}
      \end{minipage} &
      \begin{minipage}[t]{0.5\hsize}
        \centering
        \includegraphics[keepaspectratio, scale=0.25]{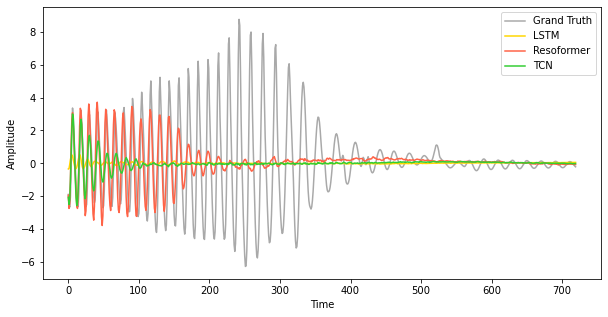} \\
        \subcaption{}
        \label{res1_b}
      \end{minipage} \\
   
      \begin{minipage}[t]{0.5\hsize}
        \centering
        \includegraphics[keepaspectratio, scale=0.25]{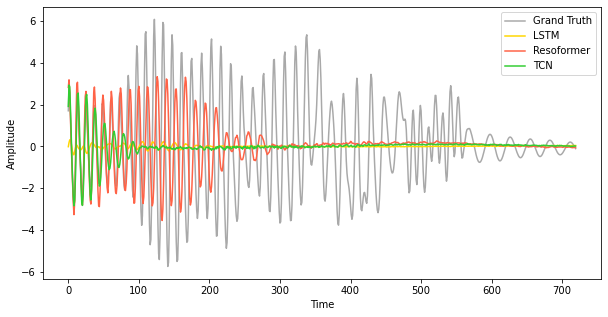} \\
        \subcaption{}
        \label{res1_c}
      \end{minipage} &
      \begin{minipage}[t]{0.5\hsize}
        \centering
        \includegraphics[keepaspectratio, scale=0.25]{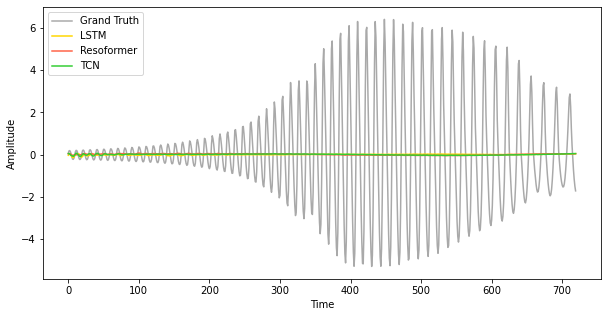} \\
        \subcaption{}
        \label{res1_d}
      \end{minipage} 
    \end{tabular}
    \caption{Forecasting results with different prediction lengths $T$ to 720 and the input length $L$ to 192.}
    \label{fig:res_t720}
\end{figure}

\begin{figure}[t]
    \begin{tabular}{cc}
      \begin{minipage}[t]{0.5\hsize}
        \centering
        \includegraphics[keepaspectratio, scale=0.25]{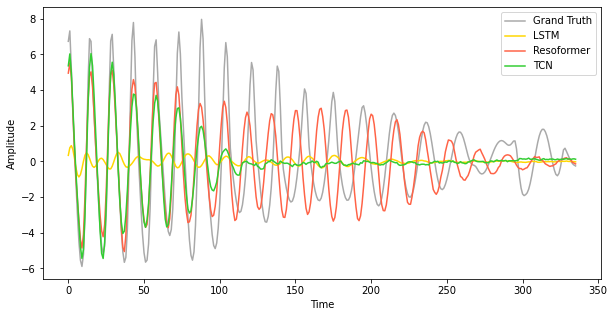} \\
        \subcaption{}
        \label{res2_a}
      \end{minipage} &
      \begin{minipage}[t]{0.5\hsize}
        \centering
        \includegraphics[keepaspectratio, scale=0.25]{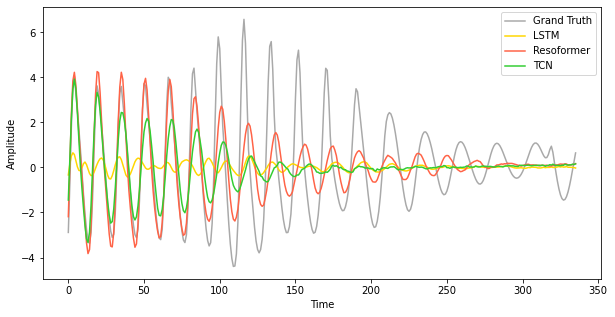} \\
        \subcaption{}
        \label{res2_b}
      \end{minipage} \\
   
      \begin{minipage}[t]{0.5\hsize}
        \centering
        \includegraphics[keepaspectratio, scale=0.25]{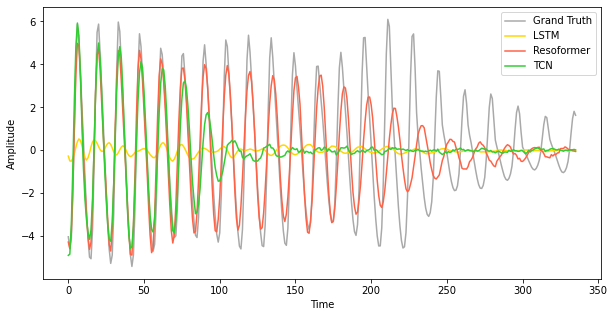} \\
        \subcaption{}
        \label{res2_c}
      \end{minipage} &
      \begin{minipage}[t]{0.5\hsize}
        \centering
        \includegraphics[keepaspectratio, scale=0.25]{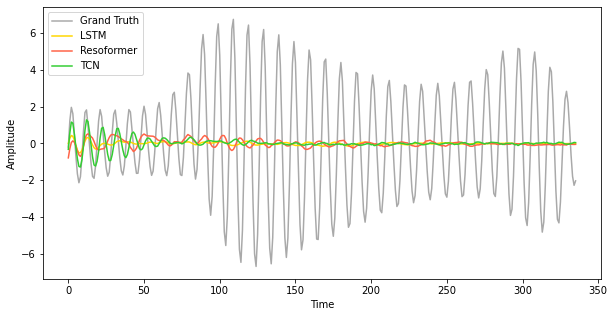} \\
        \subcaption{}
        \label{res2_d}
      \end{minipage} 
    \end{tabular}
    \caption{Forecasting results with different prediction lengths $T$ to 336 and the input length $L$ to 192.}
    \label{fig:res_t336}
  \end{figure}

In the case of medium-term forecasting ($T=336$), as visualized in~\fref{fig:res_t336}, the Resoformer's predictions closely align with the ground truth, showcasing its robustness in capturing the series' trends. However, as we extend the forecast horizon to the long term ($T=720$), depicted in~\fref{fig:res_t720}, a variation in model performance becomes evident. Despite the TCN model achieving higher scores, its limitations become apparent for distant series, with its predictions reverting to zero for longer-term forecasts—a shortfall not exhibited by the Resoformer, which consistently maintains predictive coherence.

The Resoformer's ability to sustain predictions over extended durations is credited to its innovative design, which amalgamates the short-range forecasting precision of the TCN with the LSTM's adeptness at encoding long-term temporal information. This is synthesized within the Transformer-based architecture of the Resoformer, allowing it to utilize the LSTM's long-term regression capabilities alongside the TCN's acute short-term pattern recognition.

While the figure clearly demonstrates the Resoformer's capability to track complex amplitude modulations of the ground truth waveform over time, it also reveals instances, as shown in~\fref{fig:res_t336}(d) and~\fref{fig:res_t720}(d), where long-term forecasting remains challenging, with certain oscillatory waveforms being poorly predicted.

\section{Conclusion and Future work}
\label{sec:conclusion}
In this study, we have presented a machine-learning-based damping-control technique for predicting torsional resonance in EVs using time-series data on motor rotation speed as input. Our Transformer-based model effectively captures the attention between recurrent and convolutional features extracted from the measured data points. Through extensive evaluation on a new dataset of 2,600 simulated vibration sequences, we have achieved state-of-the-art results, demonstrating the potential of our approach in enhancing torsional resonance detection and reducing excessive shaft loads.

However, we acknowledge that long-term forecasting remains a challenge, particularly in accurately predicting certain oscillatory waveforms. To address this limitation, we propose a novel approach inspired by the work of Kuniyoshi et al.~\cite{kuniyoshi2024}, which utilizes quantile regression to predict oscillation amplitudes. This method shows promise in improving the accuracy of long-term forecasts by directly targeting amplitude predictions within vibration series.

In future work, we plan to further enhance the accuracy of our model and validate its performance in real-world driving scenarios. Additionally, we aim to explore other potential applications and investigate the feasibility of integrating our technique into existing vibration-control systems.

\section*{Appendix}
In this section, we provide additional tables and figures that were obtained through our extensive benchmarking process. These supplementary materials offer further insights and detailed information to complement the main findings presented in the paper.

\subsection*{Benchmarking Results by MSE}
\Tref{tab:mse} is the results of transformers when applying them to the VIBES dataset whose metrics are the Mean Squared Error (MSE). The training settings and parameters are the same as \sref{sec:results}. The MSE is given by the formula:

\begin{equation}
\text{MSE} = \frac{1}{n} \sum_{i=1}^{n} (y_i - \hat{y}_i)^2
\end{equation}

where $n$ is the total number of data points, $y_i$ represents the actual value, and $\hat{y}_i$ represents the predicted value for the $i$-th data point.

\begin{table}[t]
    \caption{Forecasting results with different prediction lengths $T\in\{96,192,336,720\}$. We set the input length $L$ to 192 for VIBES. The highest and second-highest for each metric are indicated in bold and underlined, respectively. The metrics used are MSE.}
    \centering
    \begin{tabular}{lcccc}
        Model & T=720 & 336 & 192 & 96 \\
        \hline
        Transformer & \textbf{1.038} & \textbf{0.760} & \textbf{0.473} & \textbf{0.312} \\
        Autoformer & 1.497 & 1.434 & 1.090 & 0.869 \\
        DLinear & 1.322 & 1.247 & 1.174 & 1.116 \\
        NonStationary & \underline{1.063} & 0.889 & 0.591 & \underline{0.328} \\
        iTransformer & 1.203 & 1.035 & 0.799 & 0.560 \\
        Crossformer & 1.072 & \underline{0.764} & \underline{0.511} & 0.356 \\
        PatchTST & 1.167 & 0.964 & 0.652 & 0.448 \\
        Zero & 1.334 & 1.262 & 1.198 & 1.157 \\
    \end{tabular}
    \label{tab:mse}
\end{table}


\subsection*{Parameter Configuration in TSLib}

In the context of this paper, we detail the parameter settings for TSLib as employed in our experiments, which are summarized in \tref{tab:tslib}. The sequence length (\texttt{seq\_len}) is typically configured to 96, and the label length (\texttt{label\_len}) to 48, except for the ILI dataset, where it is adjusted to 18. The dimensions for the encoder input (\texttt{enc\_in}), decoder input (\texttt{dec\_in}), and output (\texttt{c\_out}) vary based on the dataset utilized.

The model dimension (\texttt{d\_model}) is computed using the formula:
\[
\texttt{d\_model} = \min\left\{\max\left\{2^{\left|\log C\right|}, d_{\min}\right\}, d_{\max}\right\}
\]
where \( C \) is the dimension of the input series, with \( d_{\min} \) and \( d_{\max} \) set to 32 and 512, respectively.

For the experiments detailed within this paper, \texttt{d\_model} was set to 128, and \texttt{d\_ff} to 512. This specific configuration was chosen to optimize the model performance for our forecasting tasks. In accordance with the configurations for the Autoformer, as well as the Transformer and Informer models, the label length is set to half the sequence length, such that \texttt{seq\_len} = \( 2 \times \) \texttt{label\_len}. Furthermore, we follow the precedent set by these models by defining \texttt{d\_ff} as four times the model dimension (\texttt{d\_ff} = \( 4 \times \) \texttt{d\_model}).

These parameter choices are critical to our experimental framework, ensuring that our models are finely adjusted to the specific requirements of the datasets, thereby facilitating robust and accurate forecasting performances.

\begin{table}[t]
    \caption{Transformer Parameter Specifications in TSlib. A checkmark indicates whether a parameter is utilized in the model configuration.}
    \centering
    \begin{tabular}{lccccccccc}
        Parameters & Default & Trans. & Auto. & DLinear & Stationary & Cross. & PatchTST & TimesNet & LightTS \\
        \hline
        \texttt{seq\_len} & 96 & $\checkmark$ & $\checkmark$ & $\checkmark$ & $\checkmark$ & $\checkmark$ & $\checkmark$ & $\checkmark$ & $\checkmark$ \\
        \texttt{label\_len} & 48 & & $\checkmark$ & & $\checkmark$ & & & $\checkmark$ &  \\
        \texttt{pred\_len} & 96 & $\checkmark$ & $\checkmark$ & $\checkmark$ & $\checkmark$ & $\checkmark$ & $\checkmark$ & $\checkmark$ & $\checkmark$ \\
        \texttt{output\_attention} & False & $\checkmark$ & $\checkmark$ & & $\checkmark$ & & $\checkmark$ & & \\
        \texttt{enc\_in} & 7 & $\checkmark$ & $\checkmark$ & $\checkmark$ & $\checkmark$ & $\checkmark$ & $\checkmark$ & $\checkmark$ & $\checkmark$ \\
        \texttt{dec\_in} & 7 & $\checkmark$ & $\checkmark$ & & $\checkmark$ & & & & \\
        \texttt{c\_out} & 7 & $\checkmark$ & $\checkmark$ & & $\checkmark$ & & & $\checkmark$ & \\
        \texttt{d\_model} & 512 & $\checkmark$ & $\checkmark$ & & $\checkmark$ & $\checkmark$ & $\checkmark$ & $\checkmark$ & $\checkmark$ \\
        \texttt{d\_ff} & 2048 & $\checkmark$ & $\checkmark$ & & $\checkmark$ & $\checkmark$ & $\checkmark$ & $\checkmark$ & \\
        \texttt{embed} & timeF & $\checkmark$ & $\checkmark$ & & $\checkmark$ & & & $\checkmark$ & \\
        \texttt{freq} & h & $\checkmark$ & $\checkmark$ & & $\checkmark$ & & & $\checkmark$ & \\
        \texttt{dropout} & 0.1 & $\checkmark$ & $\checkmark$ & $\checkmark$ & $\checkmark$ & $\checkmark$ & $\checkmark$ & $\checkmark$ & $\checkmark$ \\
        \texttt{factor} & 1 & $\checkmark$ & $\checkmark$ & & $\checkmark$ & $\checkmark$ & $\checkmark$ & & \\
        \texttt{n\_heads} & 8 & $\checkmark$ & $\checkmark$ & & $\checkmark$ & $\checkmark$ & $\checkmark$ & & \\
        \texttt{activation} & gelu & $\checkmark$ & $\checkmark$ & & $\checkmark$ & & $\checkmark$ & \\
        \texttt{e\_layers} & 2 & $\checkmark$ & $\checkmark$ & & $\checkmark$ & $\checkmark$ & $\checkmark$ & $\checkmark$ & \\
        \texttt{d\_layers} & 1 & $\checkmark$ & $\checkmark$ & & $\checkmark$ & & & & \\
        \texttt{num\_class} & & $\checkmark$ & $\checkmark$ & $\checkmark$ & $\checkmark$ & $\checkmark$ & $\checkmark$ & & $\checkmark$ \\
        \texttt{moving\_avg} & 25 & & $\checkmark$ & & & & & & \\
        \texttt{p\_hidden\_dims} & [128, 128] & & & & $\checkmark$ & & & & \\
        \texttt{p\_hidden\_layers} & 2 & & & & $\checkmark$ & & & & \\
        \texttt{top\_k} & 5 & & & & & & & $\checkmark$ & \\
        \texttt{num\_kernels} & 6 & & & & & & & $\checkmark$ & \\
    \end{tabular}
    \label{tab:tslib}
\end{table}

\subsection*{Hyper Parameter Tuning}
We conduct hyper parameter tuning for choosing learning rate. Optimized parameters listened in \tref{tab:optimized_lr} and parameters during 5 times training. Detailed is described in \Sref{sec:training}.

\begin{table}[t]
\begin{center}
\caption{Optimized learning rates.}
\label{tab:optimized_lr}
    \begin{tabular}{lc}
        Models & Learning Rate \\ \hline
        Transformer & $10^{-3}$ \\
        Autoformer & $10^{-4}$ \\
        DLinear & $10^{-4}$ \\
        NonStationary & $10^{-3}$ \\
        iTransformer & $10^{-4}$ \\
        Crossformer & $10^{-4}$ \\
        PatchTST & $10^{-4}$ \\
    \end{tabular}
\end{center}
\end{table}

\section*{Transformer Execution Time Measurement}
In this section, our focus is on measuring the execution time of transformers in the context of real-time in-vehicle applications. Given the time-sensitive nature of automotive systems, it is crucial to evaluate the performance of transformers in terms of their execution time.

For our measurement, we specifically set the \texttt{label\_len} parameter to 1. This choice allows us to assess the execution time when predicting a single future value based on the input sequence. Additionally, we employed the following conditions: a minibatch size of 32, a maximum of 10 training epochs, and 10,900 steps per epoch. To conduct our measurements, we utilized the NVIDIA GeForce RTX 3090 GPU.

\begin{table}[t]
\begin{center}
\caption{Transformer excution time measurement.}
\label{tab:timemeasure}
    \begin{tabular}{lcc}
        Models & Duration & Duration a epoch \\ \hline
        Transformer & 3.3 h & 20 min \\
        Autoformer & 8.8 h & 53 min \\
        DLinear & 12.5 min & 1.3 min \\
        NonStationary & 3.5 h & 21 min \\
        iTransformer & 18.2 min & 1.8 min \\
        Crossformer & 1.9 h & 11 min \\
        PatchTST & 24.8 min & 2.5 min \\
        TimesNet & N/A & N/A \\
        LightTS & 14.3 min & 1.4 min \\
    \end{tabular}
\end{center}
\end{table}

\begin{figure}[t]
    \centering
    \includegraphics[keepaspectratio, scale=0.15]{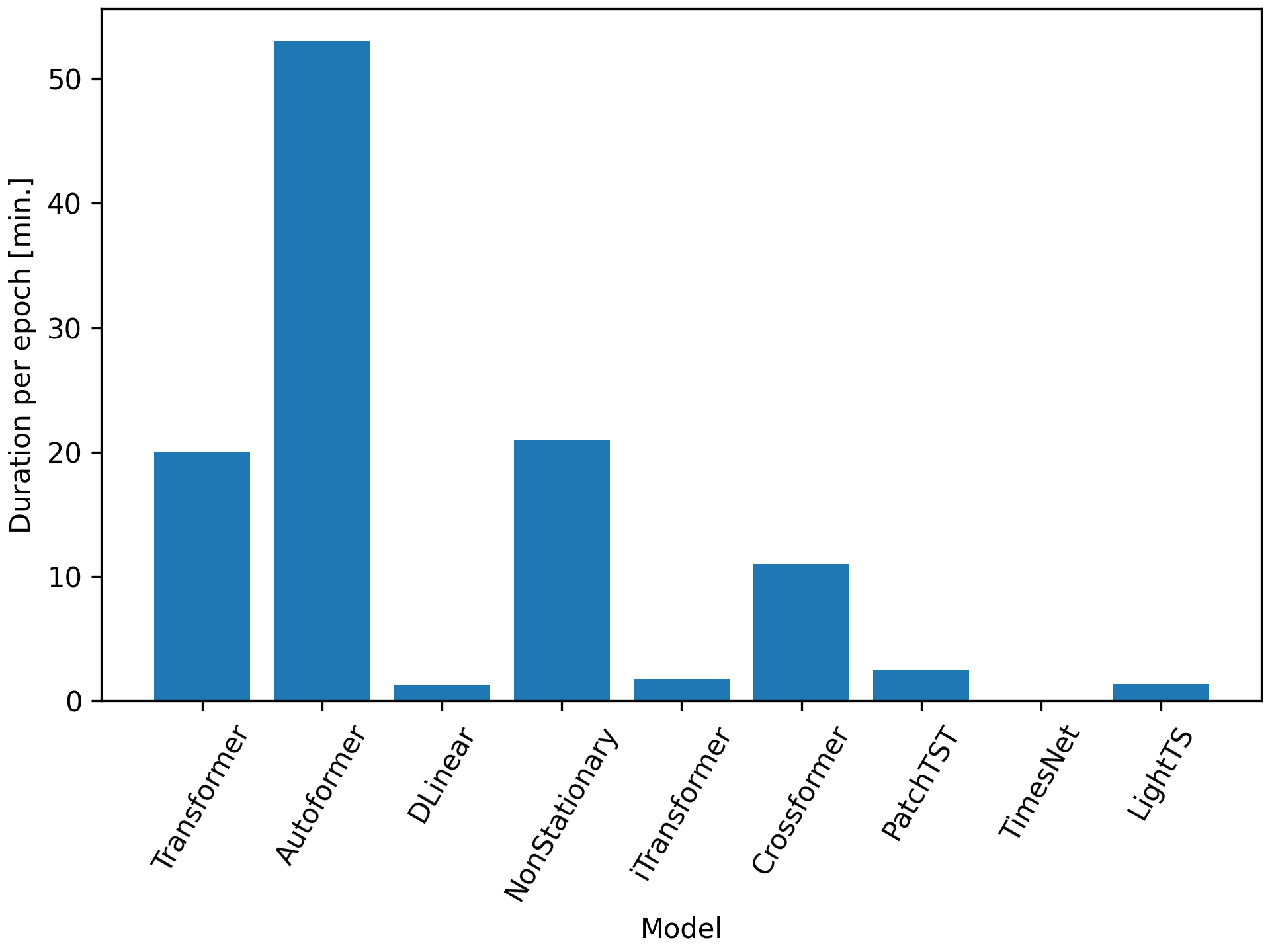}
    \caption{Transformer excution time measurement.}
    \label{fig:timemeasure}
\end{figure}

\bibliographystyle{splncs04}

\end{document}